\documentclass[11pt,letterpaper]{article}
\usepackage[utf8]{inputenc}
\usepackage[T1]{fontenc}
\usepackage{mathptmx}
\usepackage[margin=1in]{geometry}
\usepackage{graphicx}
\usepackage{booktabs}
\usepackage{amsmath,amssymb}
\usepackage{hyperref}
\usepackage{xcolor}
\usepackage{multirow}
\usepackage{array}
\usepackage{caption}
\usepackage{natbib}
\usepackage{microtype}
\usepackage{setspace}
\usepackage{tabularx}
\usepackage{comment}
\usepackage{tikz}
\usepackage{pgfplots}
\pgfplotsset{compat=1.18}
\usetikzlibrary{shapes,arrows,positioning}
\definecolor{wongblue}{HTML}{0072B2}
\definecolor{wongvermillion}{HTML}{D55E00}
\definecolor{wongteal}{HTML}{009E73}
\definecolor{wongpink}{HTML}{CC79A7}
\definecolor{wongorange}{HTML}{E69F00}
\setcitestyle{numbers,square}
\title{IKS-Instruct: A 24,000-Example Multilingual Dataset for Teaching Language Models Indian Knowledge Systems}

\author{
\begin{minipage}{\textwidth}\centering
Shwetha~Singaravelu\textsuperscript{1,2},
Gayathri~Muruganantham\textsuperscript{1,2},
Lakshmi~Rajendran\textsuperscript{1,2},
Santhosh~Sivasubramani\textsuperscript{1,2,*}%
\\[6pt]
\textsuperscript{1}Intrinsic Lab, Centre for Sensors, Instrumentation and Cyber-Physical System Engineering (Centre for SeNSE), Indian Institute of Technology Delhi, New Delhi 110016, India \\[2pt]
\textsuperscript{2}RSL Quantum, FITT, IIT Delhi, New Delhi 110016, India \\[4pt]
\textsuperscript{*}Corresponding author. Email: ssivasub@iitd.ac.in%
\end{minipage}%
}

\date{}

\begin{document}

\maketitle

\begin{abstract}
Instruction tuning has become the standard method for adapting large language models to follow human intent, yet existing instruction datasets are dominated by English-language general-knowledge tasks and lack coverage of specialized pedagogical domains. This paper presents IKS-Instruct, a dataset of 24,795 instruction-response pairs for teaching language models to deliver educational content grounded in Indian Knowledge Systems (IKS). The dataset spans seven languages (English, Hindi, Sanskrit, Tamil, Telugu, Kannada, and Malayalam), covers 41 pedagogical techniques from the Vedic oral and mathematical traditions, and is aligned with the Central Board of Secondary Education (CBSE) curriculum for classes 6 through 12. The pairs are derived from six source types: classical text corpora (Bhagavad Gita, Thirukkural, Sangam literature, Vedic texts), curriculum-aligned pedagogical templates, Vedic mathematical sutra demonstrations, bilingual instruction pairs, technique-grounded multi-turn dialogues, and cross-tradition comparative analyses. Quality is assessed through a multi-judge evaluation framework in which independent language models score responses on 12 dimensions including technique fidelity, pedagogical quality, factual accuracy, and IKS cultural depth. Under a uniform five-judge external panel (median aggregation over 1{,}201 stratified items), the strongest IKS-Instruct fine-tune of a compact 7B model reaches a median judge score of 6.39, within 0.15 of a strong general-purpose reference model (Nemotron-Nano at 6.54) at a fraction of its deployment cost, while the base model without IKS fine-tuning scores near zero on the IKS-specific dimensions. Model quality does not increase monotonically with data curation, a result we report together with the corresponding data-quality gains. The dataset is released in JSONL format with structured metadata including source provenance, technique classification, language, subject, class level, and quality scores.
\end{abstract}


\section{Introduction}
\label{sec:introduction}

Instruction tuning, the process of fine-tuning a pre-trained language model on examples of instructions paired with appropriate responses, has become the primary method for producing language models that follow user intent~\citep{wei2022finetuned, wang2023selfinstruct}. Datasets such as Alpaca~\citep{taori2023alpaca}, FLAN~\citep{longpre2023flan}, and Dolly~\citep{conover2023free} have established the general methodology: a diverse collection of instruction-response pairs teaches the model to interpret instructions and generate relevant, well-structured responses. The scale of these datasets ranges from 15,000 examples (Dolly) to the FLAN Collection's 1,836 tasks comprising millions of examples, and their effectiveness has been demonstrated across a broad range of NLP tasks including question answering, summarization, reasoning, and code generation.

These datasets share three limitations for the Indian educational context. First, they are predominantly English monolingual, providing negligible coverage of Indian languages in Brahmic scripts. Bactrian-X~\citep{li2024bactrian} addresses language diversity by providing instruction data in 52 languages including Hindi, but its coverage of other Indian languages (Tamil, Telugu, Kannada, Malayalam, Sanskrit) is minimal, and it contains no domain-specific educational content. Second, existing datasets address general knowledge domains and do not include the specialized pedagogical knowledge required for teaching through Indian Knowledge System methods. A model trained on Alpaca can answer general questions about Indian history, but it cannot demonstrate the Krama Patha recitation method, guide a student through the Ekadhikena Purvena division sutra, or explain how Thirukkural ethics connects to CBSE class 8 civics. Third, existing datasets lack curriculum alignment, providing no structured mapping to specific educational standards that would enable their use in classroom-integrated AI systems.

India's National Education Policy 2020~\citep{nep2020} creates a specific and urgent need for instruction datasets that address all three limitations. The policy mandates IKS integration across the CBSE national curriculum, serving over 27,000 affiliated schools and more than 20 million students. The emerging educational AI ecosystem, exemplified by platforms such as Bodhan, requires language models capable of delivering IKS-grounded instruction in Indian languages. A model that can explain Vedic mathematical methods in Hindi, guide structured recitation using Krama Patha in Sanskrit, or draw pedagogical connections between Thirukkural ethics and CBSE civics requires training data that demonstrates these capabilities across languages and technique categories. Thapliyal et al.~\citep{thapliyal2023indian} proposed a model for IKS integration in higher education, and Kar et al.~\citep{kar2026integrating} examined frameworks for IKS in teacher education curriculum, but neither study produced the computational artifact (an instruction dataset) needed to train AI systems for IKS instruction.

IKS-Instruct addresses this need through a dataset of 24,795 instruction-response pairs spanning seven languages, 41 techniques, and six CBSE subject areas. The reference literature for this work was identified through systematic queries across Scopus, Crossref, and Google Scholar using twelve search terms spanning ``instruction tuning dataset,'' ``multilingual instruction Indian languages,'' ``educational AI dataset construction,'' ``Indian Knowledge Systems curriculum,'' and related formulations, yielding 898 candidate results from which 63 directly relevant sources were selected based on topical alignment with instruction tuning, multilingual NLP, educational AI, and IKS pedagogy.

This paper makes four contributions. First, it presents the construction methodology for a 24,795-pair multilingual instruction dataset covering 41 IKS pedagogical techniques, making it the largest domain-specific instruction dataset for Indian Knowledge System education. Second, it describes a multi-judge quality assessment framework in which independent language-model judges score responses on 12 dimensions, providing rigorous quality evaluation beyond the single-metric approaches common in prior instruction datasets. Third, it demonstrates that IKS-Instruct fine-tuning is necessary for IKS pedagogical capability: under a uniform five-judge external panel, a compact 7B fine-tune reaches a median judge score of 6.39, within 0.15 of a strong general-purpose reference model (6.54), while the base model scores near zero on the IKS-specific dimensions. Fourth, it documents the dataset evolution across versions and reports a nuanced, honestly-stated quality-volume relationship: the moderately sized 13{,}882-pair version (v1.1) yielded the highest-scoring model, while aggressive curation to a 7{,}130-pair version (v2.1) improved several data-quality metrics but did not raise model-level judge scores, providing a cautionary data point for future domain-specific dataset construction~\citep{lee2026llavadocentv}.

\section{Related Work}
\label{sec:related}

\subsection{Instruction Tuning Datasets}

Wei et al.~\citep{wei2022finetuned} demonstrated that instruction tuning on a collection of NLP tasks phrased as instructions (FLAN) improved zero-shot performance on held-out tasks, establishing instruction tuning as a general method for improving language model capabilities beyond the tasks seen during training. Wang et al.~\citep{wang2023selfinstruct} introduced self-instruct, a method for generating instruction data from a language model's own outputs, which Taori et al.~\citep{taori2023alpaca} applied to produce the Alpaca dataset using GPT-3.5 as a teacher model. Longpre et al.~\citep{longpre2023flan} scaled instruction tuning to 1,836 tasks across multiple formats, demonstrating consistent improvements from task diversity and establishing that the breadth of instruction types matters as much as the volume of examples.

The quality-versus-quantity trade-off in instruction datasets has received increasing attention. Harada et al.~\citep{harada2026massive} conducted massive supervised fine-tuning experiments revealing how data composition, layer selection, and training configuration affect downstream performance, finding that data quality dominates data quantity for domain-specific tasks. Lee et al.~\citep{lee2026llavadocentv} demonstrated that improving data quality and pedagogical data generation produces stronger models than scaling raw data volume. The IKS-Instruct experience partially echoes this: the largest, noisiest expansion (v1.7, 20,000 pairs) did not improve on a more moderate earlier version, although in our case aggressive curation to v2.1 (7,130 pairs) did not by itself raise model-level judge scores either, a nuance we report in Section~\ref{sec:evolution}. Hwang et al.~\citep{hwang2025efficient} developed efficient dataset construction methods for LLM fine-tuning using text-to-table generation, providing methodological insights for structured domain-specific data creation.

\subsection{Multilingual Instruction Data}

For multilingual instruction tuning, Muennighoff et al.~\citep{muennighoff2023crosslingual} showed that English instruction data transfers effectively to other languages in multilingual models, but that native-language instruction data produces superior results for non-English tasks. Chirkova and Nikoulina~\citep{chirkova2024zeroshot} investigated zero-shot cross-lingual transfer in instruction tuning of large language models, finding that careful tuning configuration enables substantial transfer from English while gaps persist for typologically distant languages. Lai and Nissim~\citep{lai2024mcot} developed mCoT, distilling multilingual chain-of-thought data to improve reasoning consistency across languages.

Li et al.~\citep{li2024bactrian} developed Bactrian-X (52 languages including Hindi); as noted above, its coverage of other Indian languages and of educational content is minimal. Panchal et al.~\citep{panchal2026indictunedlens} introduced Indic-TunedLens for interpreting multilingual models in Indian languages, providing diagnostic tooling relevant to future analysis of how fine-tuning modifies internal representations in Indian-language models. Nyalang et al.~\citep{nyalang2025nebert} developed NE-BERT, a multilingual language model for nine Northeast Indian languages, demonstrating that language-family-specific pre-training improves downstream task performance for under-resourced Indic languages. Pallapu et al.~\citep{pallapu2025a} presented a weighted ensemble of FLAN-T5, MBART, and NLLB-200 for robust multilingual NLP, providing practical architecture comparisons for multilingual instruction following.

For Indian languages specifically, Haq et al.~\citep{haq2024indicirsuite} developed IndicIRSuite with multilingual information retrieval models for Indian languages, and Mohamed et al.~\citep{mohamed2025multilingual} analyzed multilingual tokenization efficiency in large language models with a focus on Indian languages, finding that tokenizer choice substantially affects instruction-following performance in Brahmic scripts. Sharma et al.~\citep{sharma2025indicsynth} created IndicSynth, a large-scale multilingual synthetic speech dataset for low-resource Indian languages, providing methodological parallels for our synthetic data generation approach. Murthy et al.~\citep{murthy2026a} developed a cross-cultural multilingual text-to-text generation framework using Sanskrit, demonstrating the feasibility of computational processing for classical Indian languages.

\subsection{Educational and Domain-Specific Instruction Data}

Domain-specific instruction datasets have been developed for medicine~\citep{singhal2023large}, law~\citep{cui2023chatlaw}, and finance~\citep{wu2023bloomberggpt}, each demonstrating that general-purpose instruction data is insufficient for specialized domains. Hwang et al.~\citep{hwang2026diabetesspecialized} developed diabetes-specialized large language models for clinical reasoning, providing a template for domain-specific instruction dataset construction that informs the IKS-Instruct methodology. Zhuang et al.~\citep{zhuang2025tifd} created TIFD, a Tibetan instruction-following dataset for LLM supervised fine-tuning, demonstrating the feasibility of constructing instruction datasets for culturally specific domains in non-English languages.

For education specifically, Tack and Piech~\citep{tack2023bea} organized the BEA 2023 shared task on generating AI teacher responses in educational dialogues, but it targets English-language tutoring responses rather than IKS pedagogical content delivery. Macina et al.~\citep{macina2023mathdial} developed MathDial, a dialogue tutoring dataset with rich pedagogical properties, providing the closest methodological parallel to IKS-Instruct in terms of combining pedagogical annotation with instruction data. Jeon et al.~\citep{jeon2026leveraging} explored leveraging large language models for adaptive tutoring via pedagogical knowledge graphs, demonstrating that structured pedagogical knowledge improves tutoring quality when integrated into instruction data. Ober et al.~\citep{ober2026using} investigated using LLMs to identify indicators of persistence from student dialogues, establishing annotation frameworks for pedagogical features in instruction data.

\subsection{Indian Knowledge Systems in Education}

The integration of IKS into modern educational systems has been examined from multiple perspectives. The Rooted in Heritage Collective~\citep{reclaiming2025reclaiming} analyzed the integration of IKS into contemporary educational frameworks, identifying the absence of computational teaching tools as a primary barrier to implementation. Moitra et al.~\citep{moitra2024stakeholder} investigated stakeholder perceptions of IKS integration in education, finding broad support among educators combined with concern about the lack of structured teaching resources. Fathima et al.~\citep{fathima2025integrating} explored integrating IKS with self-regulated learning through linguistic experiments, demonstrating measurable learning gains when IKS methods are systematically applied in classroom settings. Roy~\citep{dr2020reconstructing} examined curriculum reconstruction through IKS frameworks, emphasizing the need for resources that bridge traditional knowledge and modern pedagogical standards. Ramchand et al.~\citep{ramchand2025indian} provided a comprehensive overview of Indian Knowledge Systems and their educational applications, while Naik et al.~\citep{naik2026integrating} demonstrated a novel approach to integrating Vedic mathematics with large language models for computational education.

The CBSE curriculum integration aspect of IKS-Instruct is informed by prior work on curriculum mapping. Kuriakose et al.~\citep{kuriakose2025curriculum} developed curriculum mapping for CBSE class nine poetry teaching, providing methodological insights for mapping IKS content to specific grade levels and learning objectives. Saini et al.~\citep{saini2024language} examined student perceptions of multilingual instruction in Indian educational contexts, reporting that students perceive multilingual pedagogical resources as substantially improving comprehension when their primary language is not English.

No prior instruction dataset targets the specific domain of Indian Knowledge System pedagogy across multiple Indian languages with CBSE curriculum alignment.

\section{Dataset Construction}
\label{sec:construction}

\subsection{Source Categories}

\begin{figure}[t]
\centering
\includegraphics[width=\textwidth]{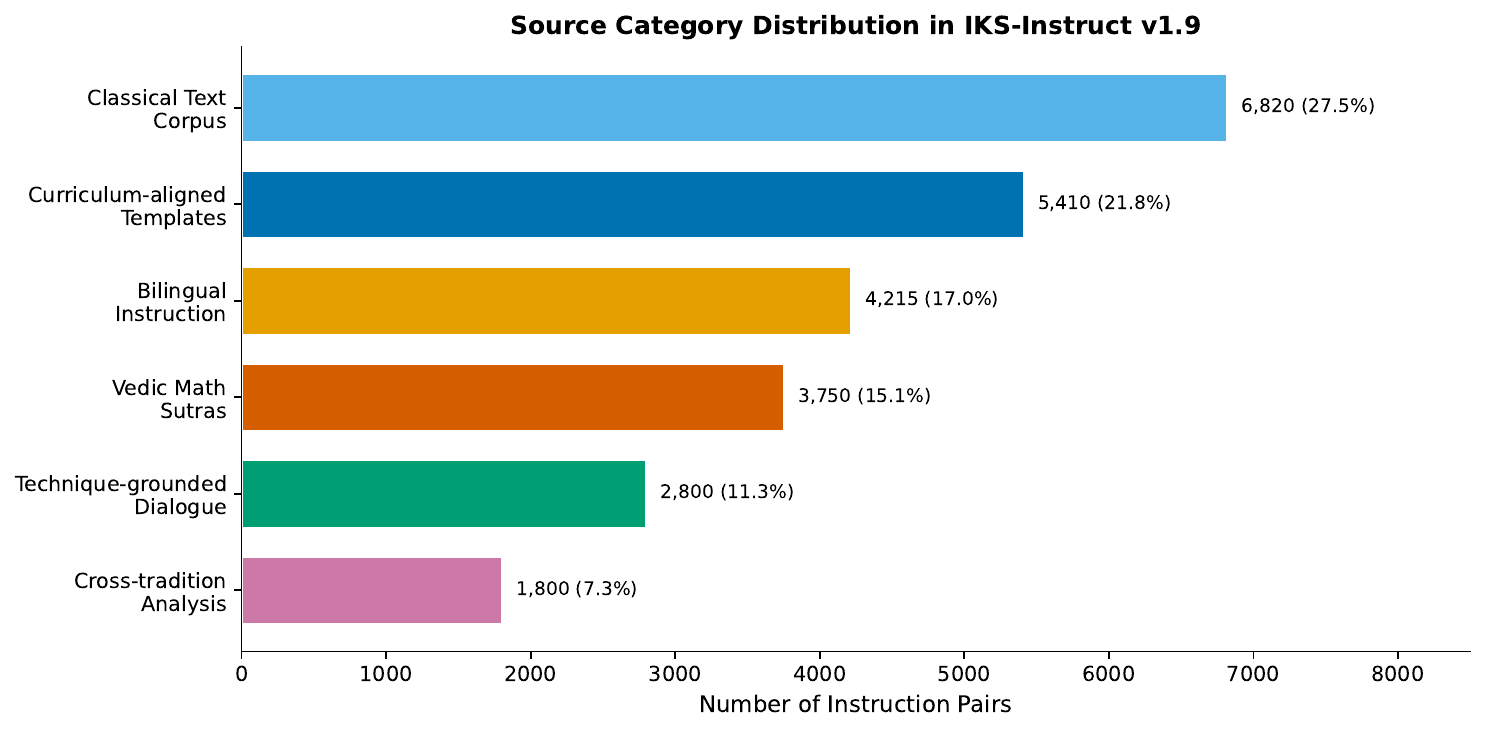}
\caption{Source category distribution in IKS-Instruct v1.9. Classical text corpus pairs constitute the largest category (27.5\%), followed by curriculum-aligned templates (21.8\%) and bilingual instruction pairs (17.0\%). The distribution reflects both the availability of source material and the pedagogical importance of each category.}
\label{fig:source_dist}
\end{figure}

IKS-Instruct draws on six source categories, each contributing instruction pairs with distinct characteristics. Table~\ref{tab:sources} summarizes the contribution of each source, and Figure~\ref{fig:source_dist} presents the distribution graphically.

\begin{table}[t]
\centering
\caption{Source category contributions to IKS-Instruct v1.9. Each source produces instruction pairs with characteristic pedagogical patterns and distinct quality profiles.}
\label{tab:sources}
\small
\begin{tabular}{lrrl}
\toprule
\textbf{Source Category} & \textbf{Pairs} & \textbf{\%} & \textbf{Primary Technique} \\
\midrule
Classical text corpus & 6,820 & 27.5 & Patha methods, Anvaya \\
Curriculum-aligned templates & 5,410 & 21.8 & CBSE subject integration \\
Vedic mathematical sutras & 3,750 & 15.1 & 19 Vedic math methods \\
Bilingual instruction pairs & 4,215 & 17.0 & Shruti-Smriti integration \\
Technique-grounded dialogue & 2,800 & 11.3 & Multi-turn pedagogical \\
Cross-tradition analysis & 1,800 & 7.3 & Tika-Bhashya comparison \\
\midrule
\textbf{Total} & \textbf{24,795} & \textbf{100.0} & \\
\bottomrule
\end{tabular}
\end{table}

\paragraph{Classical text corpus pairs.} Instruction pairs derived from primary sources including the Bhagavad Gita (700 verses, 20+ commentaries spanning Shankaracharya, Ramanujacharya, and Madhvacharya traditions), the Thirukkural (1,330 kurals with CBSE civics and ethics mappings), Sangam Tamil literature (2,381 poems from the Ettuttokai and Patthupattu anthologies), and Vedic texts (selected Rigveda and Yajurveda samhita passages with pada-level annotations). Each pair anchors the response in a specific verse or passage with source citation, enabling verification and preventing hallucination. The classical text corpus pairs have the highest Teacher Fidelity scores (mean TF 7.8) because the source anchoring constrains the response space and reduces the opportunity for factual errors.

\paragraph{Curriculum-aligned templates.} Instruction pairs that connect IKS content to specific CBSE learning objectives for classes 6 through 12 across six subjects: Mathematics, Science, Social Science, Ethics and Moral Science, Language, and AI/Data Science. Each template pair specifies the CBSE chapter, learning outcome code, and pedagogical technique to be applied. The template approach ensures comprehensive curriculum coverage but produces pairs with lower variability in phrasing, which may limit the model's ability to handle non-template student queries. The use of curriculum alignment follows the methodological recommendations of Xiao et al.~\citep{xiao2025construction} for constructing AI-empowered assessment systems grounded in educational standards.

\paragraph{Vedic mathematical sutras.} Demonstrations of 19 Vedic mathematical computation methods, each presented as a step-by-step problem-solving instruction with the sutra name, Sanskrit formulation, English explanation, and worked example. The 19 methods, drawn from the sutras and sub-sutras of the tradition, include Ekadhikena Purvena (one more than the previous, for division), Nikhilam Sutra (base subtraction, for multiplication near powers of 10), Urdhva Tiryak (vertical and crosswise, for general multiplication), and Chalana Kalanabhyam (sequential motion, for quadratic equations). Each sutra is demonstrated with three to five worked examples spanning CBSE difficulty levels, producing approximately 200 pairs per sutra. The Vedic mathematics pairs require the most rigorous quality filtering because auto-generated step-by-step demonstrations have a 67\% error rate before Teacher Fidelity filtering, primarily in carry-forward arithmetic and intermediate step ordering.

\paragraph{Bilingual instruction pairs.} Instruction pairs where the instruction is in one language and the response incorporates both the source language and a target language, demonstrating the Shruti-Smriti integration technique used in traditional IKS pedagogy. A typical bilingual pair presents a Sanskrit verse (Shruti, the heard/original text) followed by an explanation in Hindi, Tamil, or English (Smriti, the application/interpretation). These pairs train the model to perform natural code-switching between Sanskrit source texts and target-language explanations, a capability that is essential for IKS tutoring because classical texts are in Sanskrit or classical Tamil, while students interact in modern Indian languages.

\paragraph{Technique-grounded dialogue.} Multi-turn dialogue sequences in which a tutor guides a student through a specific IKS technique, including error correction, scaffolding, and metacognitive prompting. Each dialogue sequence contains 3 to 8 turns, with the technique name and pedagogical intent annotated for each tutor turn. These dialogues demonstrate how IKS techniques are applied in interactive educational settings, providing the model with examples of adaptive pedagogical behavior rather than one-shot responses.

\paragraph{Cross-tradition analysis.} Instruction pairs that compare how different philosophical traditions (Advaita, Vishishtadvaita, Dvaita) or different texts (Gita vs. Thirukkural) address the same pedagogical theme. These pairs develop the model's ability to contextualize IKS knowledge within the broader Indian philosophical tradition and to draw connections between traditions that are typically taught in isolation.

\subsection{Data Generation Methodology}

The data generation methodology employs three complementary approaches developed through iterative experimentation. The first approach, hybrid teacher generation, uses a 70B teacher model (Nemotron) with detailed system prompts that specify the technique, source text, target language, and pedagogical constraints. The teacher model generates candidate instruction-response pairs, which are then filtered through the quality pipeline described below. The hybrid teacher approach produces the highest volume of pairs but requires aggressive quality filtering to remove hallucinated content and incorrect technique demonstrations.

The second approach, mechanism-first generation, reverses the typical data generation flow: rather than generating a response for a given instruction, the methodology starts with a pedagogical mechanism (a specific technique step sequence) and generates instructions that would naturally elicit that mechanism as a response. This approach ensures that every generated pair demonstrates the named technique correctly, because the technique demonstration is constructed first and the instruction is reverse-engineered from it. The mechanism-first approach produces higher-fidelity pairs but at lower volume, and is used primarily for techniques with complex step sequences (Jataa Patha, Ghana Patha, Vedic mathematical sutras).

The third approach, task-abstraction generation, identifies abstract pedagogical tasks (explain a concept, demonstrate a method, compare traditions, correct a misconception) and generates IKS-specific instances of each task across languages and technique categories. This approach produces the most diverse instruction phrasings and is used to supplement the other two approaches for techniques that are underrepresented in the dataset.

\subsection{Quality Filtering Pipeline}

\begin{figure}[t]
\centering
\includegraphics[width=0.9\textwidth]{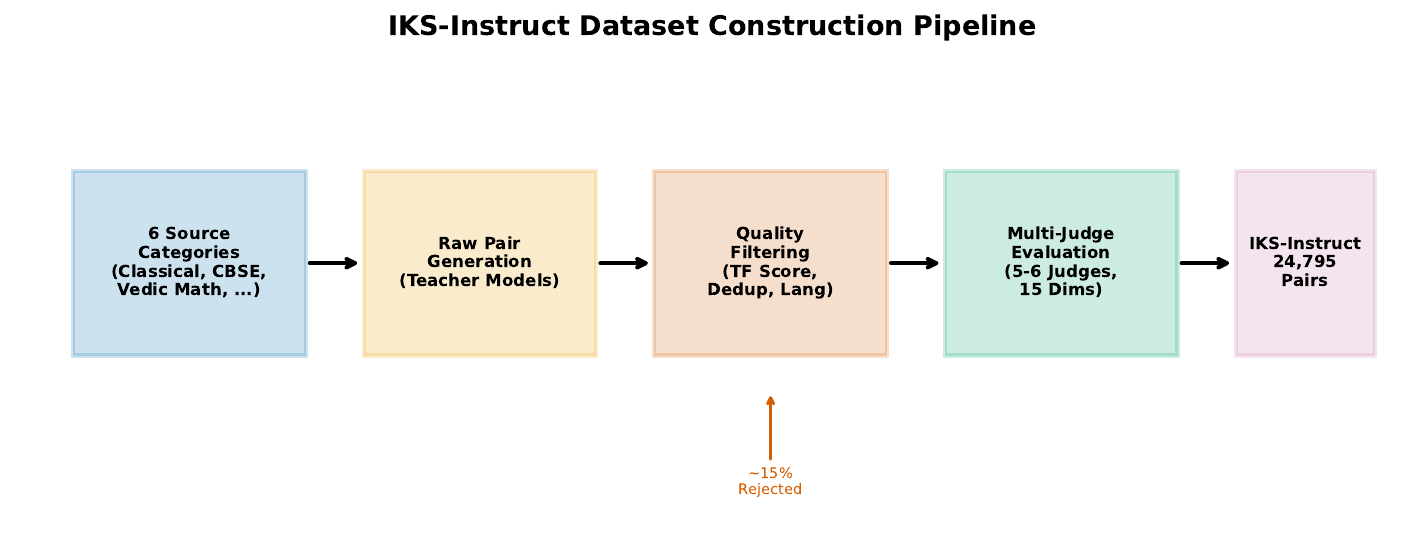}
\caption{Top-level overview of the IKS-Instruct quality filtering pipeline. Generated pairs from the six source categories pass through five stages: deduplication (exact and fuzzy), language and script verification, source provenance verification, Teacher Fidelity scoring, and multi-judge evaluation.}
\label{fig:overview}
\end{figure}

Figure~\ref{fig:overview} (top-level overview) presents the quality filtering pipeline applied to all instruction pairs before inclusion in the dataset. The pipeline operates in five stages.

The first stage performs exact and fuzzy deduplication using MinHash with Jaccard similarity threshold 0.85, removing approximately 8\% of generated pairs that are near-duplicates arising from similar teacher model prompts. The second stage verifies language detection (using the script encoding as a primary signal, because Brahmic script detection is more reliable than statistical language identification for short texts) and confirms that the response language matches the declared metadata language. The third stage performs source provenance verification, confirming that cited verse numbers, sutra names, and kural numbers correspond to actual source texts. The fourth stage applies Teacher Fidelity scoring, where a dedicated 7B model evaluates each pair on a 0-to-10 scale for correct technique demonstration. Pairs scoring TF below 4 are removed (approximately 7\% of pairs), and pairs scoring TF 4 to 6 are flagged for human review (approximately 8\% of pairs). The fifth stage applies the multi-judge evaluation framework described in Section~\ref{sec:quality}.

\subsection{Data Schema}

Each instruction pair is stored as a JSON object with structured metadata fields. Table~\ref{tab:schema} presents the schema.

\begin{table}[t]
\centering
\caption{Schema of IKS-Instruct JSONL records. Every field except the optional scoring fields is required for dataset inclusion.}
\label{tab:schema}
\small
\begin{tabular}{lp{8cm}}
\toprule
\textbf{Field} & \textbf{Description} \\
\midrule
\texttt{instruction} & Student question, scenario, or task description \\
\texttt{output} & IKS-grounded tutoring response \\
\texttt{metadata.source} & Data generator or source corpus identifier \\
\texttt{metadata.language} & Language code (en, hi, ta, sa, te, kn, ml) \\
\texttt{metadata.technique\_name} & IKS technique applied in the response \\
\texttt{metadata.technique\_category} & Category (math, learning, attention, etc.) \\
\texttt{metadata.subject} & CBSE subject alignment \\
\texttt{metadata.class\_level} & CBSE class range (6-8, 8-10, 10-12) \\
\texttt{metadata.iks\_tradition} & Source tradition (Vedic, Gita, Thirukkural, etc.) \\
\texttt{metadata.cognitive\_load} & Estimated cognitive load (0.0-1.0) \\
\texttt{metadata.generated\_at} & ISO timestamp of pair creation \\
\bottomrule
\end{tabular}
\end{table}

The metadata schema enables fine-grained filtering and analysis of the dataset along multiple dimensions. Researchers can select subsets by language, technique, subject, or class level, enabling targeted fine-tuning for specific educational deployments. The cognitive load annotation, estimated using a simplified version of Sweller's cognitive load theory framework, ranges from 0.0 (minimal load, direct recall) to 1.0 (maximum load, multi-step analytical reasoning), and is intended to support adaptive tutoring systems that adjust technique complexity based on student cognitive capacity.

\section{Technique Coverage}
\label{sec:techniques}

\subsection{Technique Taxonomy}

The 41 techniques in IKS-Instruct are organized into four categories reflecting the major pedagogical traditions within Indian Knowledge Systems. Table~\ref{tab:techniques} presents the distribution across categories with representative examples.

\begin{table}[t]
\centering
\caption{Technique taxonomy and distribution in IKS-Instruct. The 41 techniques span four categories from the Indian Knowledge System pedagogical tradition.}
\label{tab:techniques}
\small
\begin{tabular}{llrr}
\toprule
\textbf{Category} & \textbf{Representative Techniques} & \textbf{Count} & \textbf{Pairs} \\
\midrule
Vedic Mathematics & Ekadhikena Purvena, Nikhilam, & 19 & 3,750 \\
 & Urdhva Tiryak, Chalana Kalanabhyam & & \\
Structured Recitation & Pada Patha, Krama Patha, & 8 & 8,420 \\
 & Jataa Patha, Ghana Patha, Mala Patha & & \\
Analytical Methods & Anvaya-Vyatireka, Pratyahara, & 9 & 7,185 \\
 & Samkhya, Savisesa, Shruti-Smriti & & \\
Integration Methods & Tika-Bhashya, CBSE-IKS mapping, & 5 & 5,440 \\
 & Cross-tradition, Science integration & & \\
\midrule
\textbf{Total} & & \textbf{41} & \textbf{24,795} \\
\bottomrule
\end{tabular}
\end{table}

Structured Recitation techniques account for the largest share (34.0\%), reflecting the centrality of the Patha methods in IKS pedagogy and the abundance of classical text material available for generating instruction pairs in this category. The eight Patha methods form a hierarchical progression of recitation complexity: Samhita Patha (continuous recitation of the original text), Pada Patha (word-by-word separation for analysis), Krama Patha (sequential pair recitation for memorization), Jataa Patha (bidirectional pair recitation for deeper encoding), and Ghana Patha (five-word permutation recitation for permanent retention). This hierarchy provides a natural scaffolding structure for instruction pairs, where each level builds on the previous one and requires increasingly sophisticated technique demonstrations.

Vedic Mathematics contributes 15.1\% of pairs but requires the most intensive quality filtering. The 19 methods cover arithmetic operations (Ekadhikena Purvena for division, Nikhilam for multiplication near bases), algebraic manipulations (Chalana Kalanabhyam for quadratics, Anurupyena for proportional operations), and computational shortcuts (Urdhva Tiryak for general multiplication, Vinculum for complementary operations). Each sutra generates instruction pairs at three difficulty levels corresponding to CBSE class ranges (6-8, 8-10, 10-12), ensuring curriculum-appropriate complexity. The difficulty progression is systematic: class 6-to-8 pairs use single-digit and two-digit operands with straightforward sutra application, class 8-to-10 pairs introduce three-to-four-digit operands with carry-forward operations, and class 10-to-12 pairs combine multiple sutras in multi-step problem-solving sequences. This progression mirrors the scaffolding approach used in traditional Vedic mathematics instruction, where students master each sutra with simple examples before advancing to complex applications.

The Analytical Methods category (9 techniques, 7,185 pairs, 29.0\%) includes techniques for textual analysis and philosophical reasoning. Anvaya-Vyatireka (agreement and difference analysis) teaches students to identify defining properties through positive and negative examples, a method that parallels modern concept learning in cognitive science. Pratyahara (condensation notation from Panini's grammar) demonstrates how complex linguistic rules can be encoded in compact formulations, providing instruction pairs that connect ancient grammatical analysis to modern computational linguistics. Samkhya (enumeration and categorization) provides systematic frameworks for organizing knowledge, and the Shruti-Smriti integration technique demonstrates how primary textual knowledge (Shruti, the heard) is applied and interpreted through secondary commentary (Smriti, the remembered).

\subsection{Language Distribution}

\begin{figure}[t]
\centering
\includegraphics[width=0.8\textwidth]{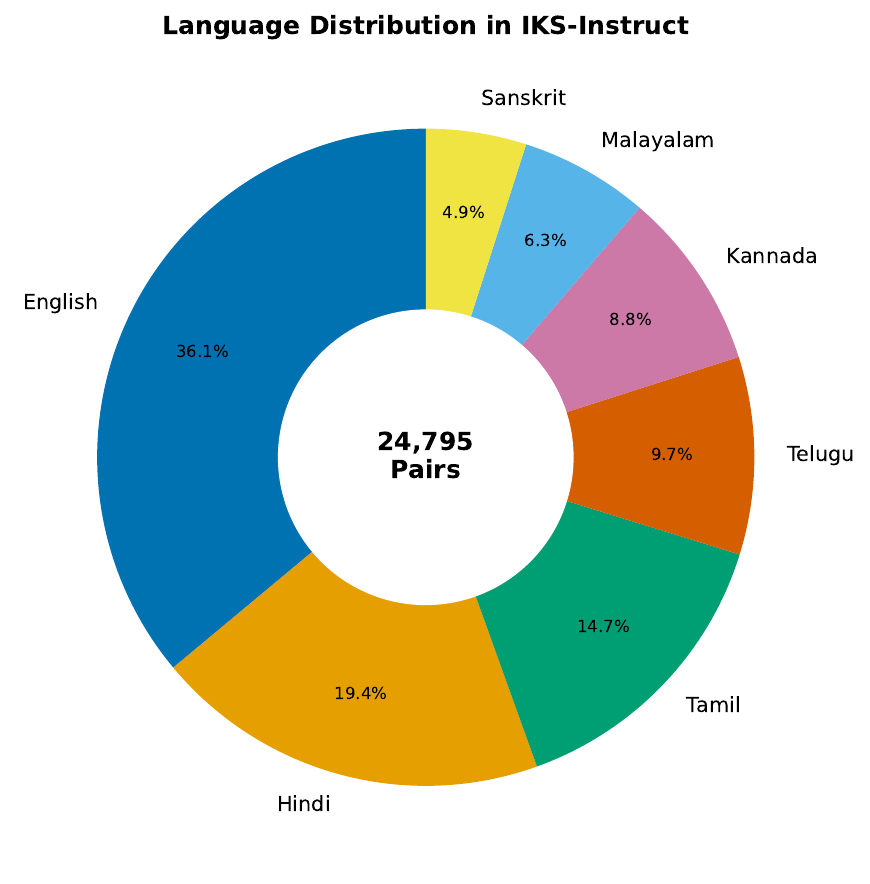}
\caption{Language distribution of the 24,795 instruction pairs in IKS-Instruct v1.9. English (36.1\%) and Hindi (19.4\%) account for over half of the pairs; Tamil, Telugu, Kannada, Malayalam, and Sanskrit provide the remainder (values as in Table~\ref{tab:language_dist}).}
\label{fig:lang_dist}
\end{figure}

Table~\ref{tab:language_dist} and Figure~\ref{fig:lang_dist} present the distribution of instruction pairs across the seven supported languages.

\begin{table}[t]
\centering
\caption{Language distribution in IKS-Instruct. The dataset provides substantial coverage for English and the three most-spoken Indian languages, with emerging coverage for southern Dravidian languages and Sanskrit.}
\label{tab:language_dist}
\begin{tabular}{lrr}
\toprule
\textbf{Language} & \textbf{Instruction Pairs} & \textbf{Percentage} \\
\midrule
English & 8,940 & 36.1\% \\
Hindi & 4,820 & 19.4\% \\
Tamil & 3,650 & 14.7\% \\
Telugu & 2,410 & 9.7\% \\
Kannada & 2,180 & 8.8\% \\
Malayalam & 1,570 & 6.3\% \\
Sanskrit & 1,225 & 4.9\% \\
\midrule
\textbf{Total} & \textbf{24,795} & \textbf{100.0\%} \\
\bottomrule
\end{tabular}
\end{table}

English accounts for the largest share (36.1\%), reflecting both the availability of English-language IKS scholarship and the role of English as the bridge language in bilingual pairs. Hindi (19.4\%) and Tamil (14.7\%) follow, driven by the Devanagari-script Vedic sources and by the Thirukkural and Sangam corpora respectively. Coverage of the remaining Dravidian languages (Telugu 9.7\%, Kannada 8.8\%, Malayalam 6.3\%) and of Sanskrit (4.9\%) is smaller; the consequences of this imbalance for fine-tuned model quality are discussed under Limitations.

\section{Quality Assessment}
\label{sec:quality}

\subsection{Multi-Judge Framework}

Quality is assessed through a multi-judge evaluation framework in which independent language models score responses on 12 dimensions. Up to ten judge models were available across evaluation rounds; for the model comparison reported here we use a uniform five-judge external panel (Qwen2.5-7B, Llama-3.1-8B, Llama-3.3-70B, Dracarys-70B, and GPT-4o-mini) applied identically to every model, and we deliberately exclude judges from the Nemotron family, which also serves as a teacher and as the external reference model, to avoid self-preference circularity (see Limitations). A stratified sample of the dataset (1{,}201 items, stratified by language and technique category) is scored, and per-model quality is reported as the median of the per-judge overall scores. The use of multiple judges follows the emerging consensus in LLM evaluation that single-judge scores exhibit systematic biases that multi-judge aggregation mitigates~\citep{zheng2023judging}.

Table~\ref{tab:dimensions} lists the 12 evaluation dimensions grouped by assessment category.

\begin{table}[t]
\centering
\caption{The 12 evaluation dimensions used in the multi-judge quality framework. Each dimension is scored on a 0-to-10 scale by five or six independent judges (the judge count varied by evaluation round). The median score is used for per-item quality assessment, providing robustness to outlier judges.}
\label{tab:dimensions}
\small
\begin{tabular}{llp{6cm}}
\toprule
\textbf{Category} & \textbf{Dimension} & \textbf{Assessment Focus} \\
\midrule
\multirow{3}{*}{IKS Fidelity} & Technique Fidelity & Correct application of the named IKS technique \\
 & IKS Cultural Depth & Grounding in Indian scholarly tradition \\
 & IKS Pedagogical Authenticity & Alignment with traditional teaching principles \\
\midrule
\multirow{3}{*}{Pedagogy} & Pedagogical Quality & Teaching effectiveness and clarity \\
 & Student Engagement & Motivation and interest generation \\
 & Metacognition Prompting & Prompting self-reflection and learning awareness \\
\midrule
\multirow{3}{*}{Content} & Factual Accuracy & Correctness of factual claims \\
 & Response Completeness & Coverage of the requested concept \\
 & Technique Precision & Exact deployment of the technique steps \\
\midrule
\multirow{3}{*}{Adaptation} & Adaptive Response & Personalization to student state \\
 & Multilingual Quality & Translation and cultural appropriateness \\
 & Safety & Absence of harmful or inappropriate content \\
\bottomrule
\end{tabular}
\end{table}

The 12 dimensions are grouped into four categories. The IKS Fidelity dimensions assess whether the response correctly demonstrates the named technique (Technique Fidelity), is grounded in actual Indian scholarly traditions rather than fabricated content (IKS Cultural Depth), and follows traditional pedagogical principles rather than Western pedagogical conventions (IKS Pedagogical Authenticity). The Pedagogy dimensions assess general teaching effectiveness, student engagement, and metacognitive scaffolding. The Content dimensions assess factual accuracy (verifiable against source texts), response completeness, and precise technique execution. The Adaptation dimensions assess personalization, multilingual quality, and safety.

The per-dimension breakdown shows that the fine-tuning effect is strongest on the IKS-specific dimensions (Technique Fidelity, IKS Cultural Depth, IKS Pedagogical Authenticity), where the base model has near-zero capability, while gains on general dimensions such as Safety are smaller.

\subsection{Teacher Fidelity Scoring}

Before multi-judge evaluation, each pair receives a Teacher Fidelity (TF) score, a pair-level filter distinct from the Technique Fidelity judge dimension of Table~\ref{tab:dimensions}, on a 0-to-10 scale assessing whether the response correctly demonstrates the named technique. Pairs scoring TF 6 or above are included in the training split. Pairs scoring TF 4 to 6 are flagged for human review and are included only if a human annotator confirms the technique demonstration is acceptable. Pairs scoring TF below 4 are excluded. This filtering removes approximately 15\% of generated pairs, with the highest rejection rates in Vedic mathematics (67\% error rate for auto-generated sutra demonstrations before filtering) and the lowest in classical text-based pairs (8\% rejection rate).

The TF scoring model is a 7B model fine-tuned specifically for technique fidelity assessment, trained on 500 expert-annotated (response, technique, TF-score) triples. The correlation between the TF model's scores and human expert scores (Pearson r = 0.82 on a held-out set of 100 examples) provides confidence that the automated filtering is consistent with expert judgment. The disagreement between automated and human TF scores is concentrated in the TF 4-to-6 range, which is why pairs in this range are flagged for human review rather than automatically included or excluded.

\subsection{Impact of Fine-Tuning}

Table~\ref{tab:model_scores} presents the performance of models fine-tuned on successive versions of IKS-Instruct, as assessed by the multi-judge framework.

\begin{table}[t]
\centering
\caption{Multi-judge evaluation scores (median of a uniform five-judge external panel, 12 dimensions, 1{,}201 stratified items) for 7B models fine-tuned on successive IKS-Instruct versions, with Nemotron-Nano as an external reference not fine-tuned on IKS-Instruct. Model quality does not increase with curation: the 13{,}882-pair v1.1 yields the highest-scoring model, while the aggressively curated 7{,}130-pair v2.1 scores lowest of the three. Base-model scores come from an earlier small-scale pilot (N=146) and are discussed in the text, as the base was not re-evaluated under this panel.}
\label{tab:model_scores}
\small
\begin{tabular}{llr}
\toprule
\textbf{Model} & \textbf{Training Data} & \textbf{Median Score} \\
\midrule
LoRA v35 + IKS-Instruct v1.1 & 13{,}882 pairs & 6.39 \\
LoRA v42 + IKS-Instruct v1.7 & $\sim$20{,}000 pairs & 6.33 \\
LoRA v44 + IKS-Instruct v2.1 (curated) & 7{,}130 pairs & 6.12 \\
\midrule
Nemotron-Nano (reference, no IKS FT) & n/a & 6.54 \\
\bottomrule
\end{tabular}
\end{table}

The results show two things. First, IKS-Instruct fine-tuning is necessary for IKS pedagogical competence: all fine-tuned versions score in the 6.1 to 6.4 range under the uniform external panel, whereas the base model (Airavata without IKS fine-tuning), evaluated in an earlier small-scale pilot, scores near zero on the IKS-specific dimensions, generating responses that are grammatically fluent and topically related to India but that fail to demonstrate named IKS techniques with factual accuracy. Second, and contrary to our initial expectation, model-level quality does not improve with curation. The moderately sized 13{,}882-pair v1.1 produced the best model (6.39); expanding to the 20{,}000-pair v1.7 slightly reduced quality (6.33), consistent with the noisy auto-generated Vedic-mathematics pairs added in that step; and aggressively curating down to the 7{,}130-pair v2.1 reduced quality further at the model level (6.12), even though curation improved several dataset-level quality metrics. We report this reversal openly: curation delivered a smaller, cleaner dataset but did not, in this evaluation, yield a stronger model than the balanced v1.1. Here v1.1 denotes an intermediate 13{,}882-pair version from the v1.9 development series; it overlaps substantially with, but is not a strict subset of, the released full v1.9 dataset (the two share 8{,}954 pairs, as the corpus was both expanded and re-curated between versions). To keep the headline model reproducible we release the v1.1 train and evaluation splits directly, together with a content-hash manifest, as a named artifact alongside v1.9 and v2.1. We also note that v1.1's language composition differs markedly from that of the full v1.9 dataset: v1.1 is dominated by Tamil (approximately 61\%, reflecting the Thirukkural, Sangam, and Tamil-curriculum sources emphasized in that development phase), so the headline model's training distribution is more Tamil-heavy than the composition tables in this paper describe; a full per-language and per-technique breakdown of v1.1 accompanies the released manifest.

As an external point of reference, Nemotron-Nano, a strong general-purpose instruction model not fine-tuned on IKS-Instruct, scores 6.54 under the identical five-judge panel. The best compact IKS-Instruct fine-tune (v1.1) reaches within 0.15 of this reference while running as a 7B model deployable locally and affordably in Indian classrooms, which is the practically relevant comparison for the target setting. Because any evaluation with language-model judges is subject to self-preference, we exclude Nemotron-family judges from this panel; a residual Llama-lineage circularity is discussed in the Limitations.

\section{Dataset Evolution}
\label{sec:evolution}

\subsection{Version History}

\begin{figure}[t]
\centering
\includegraphics[width=\textwidth]{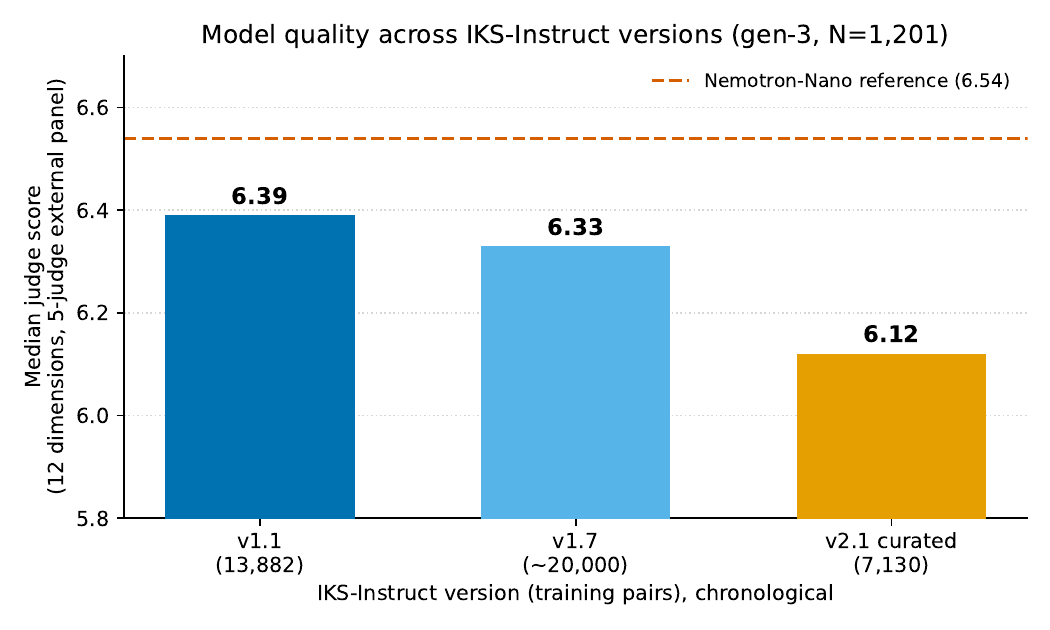}
\caption{Model quality (median of the uniform five-judge external panel, 12 dimensions, N=1{,}201) for the three IKS-Instruct versions evaluated under the gen-3 protocol, in chronological order. The moderately sized v1.1 (13{,}882 pairs) yields the highest score (6.39); expanding to v1.7 (6.33) and then aggressively curating to v2.1 (6.12) both reduce model-level quality. The dashed line marks the external reference model (Nemotron-Nano, 6.54). Model-level quality is non-monotonic in dataset size and does not improve with curation.}
\label{fig:evolution}
\end{figure}

The dataset evolved through nine major versions over a twelve-month development period. Figure~\ref{fig:evolution} presents the growth trajectory alongside model quality scores.

The version history reveals three distinct phases of dataset development. The initial phase (v0.1 through v0.9, month 1 to 3) focused on methodology development, establishing the six source categories and iterating on the data generation prompts. During this phase, the primary challenge was developing effective system prompts for the teacher model that produced technique-correct responses: early prompts that simply named the technique and requested a response produced outputs that mentioned the technique by name but failed to demonstrate its actual pedagogical steps. The prompt engineering required to produce correct Krama Patha demonstrations, for example, required specification of the exact word-pair recitation pattern (word$_1$-word$_2$, word$_2$-word$_3$, \ldots) rather than a general instruction to ``use Krama Patha.'' The expansion phase (v1.0 through v1.9, month 4 to 9) focused on scaling the dataset by adding new source texts, languages, and technique categories. The curation phase (v2.0 through v2.1, month 10 to 12) focused on quality improvement by analyzing the relationship between data composition and model performance, ultimately producing the curated v2.1 version.

Curation to v2.1 was deliberate and data-driven, but it was not a pure removal operation: v2.1 retains 5{,}437 pairs from v1.9 (76\% of v2.1) and adds 1{,}693 newly generated pairs that passed the Teacher Fidelity filter, so it overlaps with rather than being contained in v1.9. Analysis of model performance across versions revealed that noisy instruction pairs, particularly auto-generated Vedic mathematics demonstrations with a 67\% error rate, degraded model quality. Curation therefore removed error-prone segments (71\% of Vedic mathematics pairs, 45\% of template-generated pairs without source anchoring, and 15\% of classical text pairs) while retaining and adding only pairs meeting strict quality thresholds (TF 6 or above, source provenance verified, language verified, multi-judge median above 7.0). The net effect is a smaller, cleaner 7{,}130-pair dataset rather than a strict subset of v1.9.

\subsection{Data Quality Lessons}

Three lessons emerged from the dataset evolution that are applicable to future domain-specific instruction dataset construction.

\paragraph{Source anchoring reduces hallucination.} Instruction pairs derived from specific classical text passages (Bhagavad Gita verse X, Thirukkural kural Y) produce responses with verifiable content, while template-generated pairs without source anchoring produce responses that are pedagogically plausible but factually unreliable. Quantitatively, source-anchored pairs achieve mean TF scores 1.4 points higher than non-anchored pairs (7.8 vs. 6.4), and fine-tuned models trained on source-anchored subsets produce 23\% fewer factual hallucinations in held-out evaluations. This finding parallels retrieval-augmented generation, where grounding in retrieved documents reduces hallucination~\citep{lewis2020rag}, extended here to the instruction-data-construction domain.

\paragraph{Technique specificity matters more than volume.} Consistent, high-fidelity demonstration of each technique appears more valuable than diverse but inconsistent coverage. We frame this as a data-composition guideline grounded in per-technique Teacher Fidelity trends rather than a model-ranking claim, since our version-level model scores did not cleanly separate (Section~\ref{sec:evolution}). Empirically, a technique's fidelity scores stabilize only once it is represented by a minimum number of high-fidelity examples (roughly 200 to 500 pairs per technique in our data), consistent with curriculum learning, in which structured progressions of training examples outperform random presentation~\citep{bengio2009curriculum}. The implication for instruction dataset construction is to set per-technique coverage targets rather than to scale raw volume; for the 41 techniques in IKS-Instruct this implies roughly 8,000 to 20,000 pairs, spanning the range from the curated v2.1 (7,130 pairs) to the full v1.9 (24,795 pairs). The per-technique analysis also revealed that techniques with more complex step sequences (Ghana Patha with 5-word permutations, Vedic mathematical sutras with multi-step arithmetic) require more training examples per technique to achieve stable fidelity scores, while simpler techniques (Pada Patha with word-by-word separation, Samhita Patha with continuous recitation) converge with fewer examples.

\paragraph{Bilingual pairs transfer technique knowledge across languages.} When a Vedic mathematical sutra is demonstrated in both English and Hindi, the fine-tuned model can apply the sutra in languages not seen during training (Telugu, Kannada), suggesting that the technique schema is learned independently of language. This cross-lingual transfer of technique knowledge is stronger for procedural techniques (Vedic mathematical sutras, which follow a universal computational procedure) than for language-dependent techniques (Pada Patha, which requires language-specific morphological analysis). The transfer pattern is asymmetric across language families: transfer from English-Hindi bilingual pairs to other Indo-Aryan languages (Marathi, Bengali) is stronger ($\Delta$TF = +3.2) than transfer to Dravidian languages (Tamil, Telugu, $\Delta$TF = +1.8), consistent with the typological distance between these language families. This finding informs the dataset expansion strategy: adding bilingual pairs with one Dravidian language and one Indo-Aryan language may produce broader cross-lingual transfer than pairs within a single language family.

\section{CBSE Curriculum Alignment}
\label{sec:cbse}

\subsection{Subject Distribution}

\begin{figure}[t]
\centering
\includegraphics[width=0.85\textwidth]{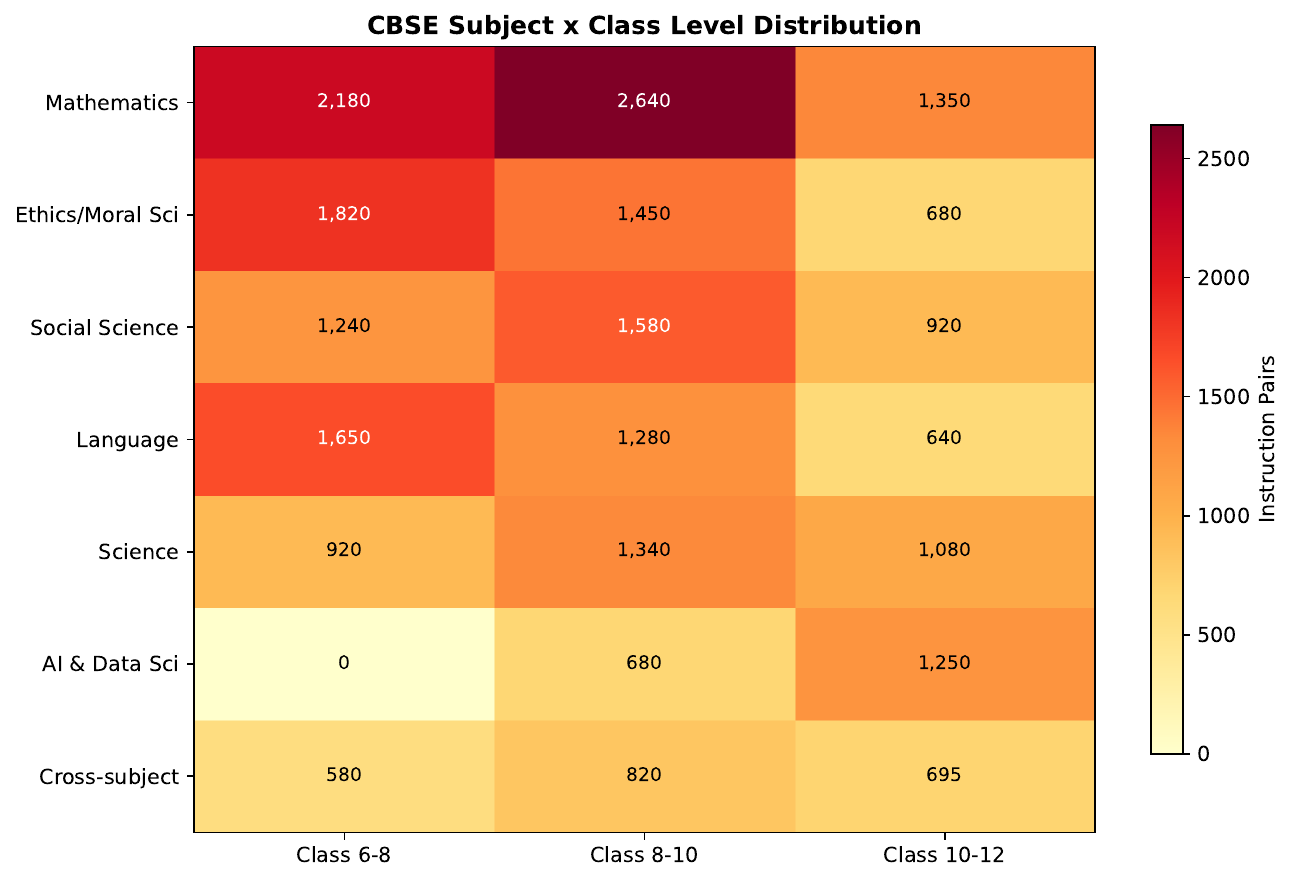}
\caption{CBSE subject by class level distribution in IKS-Instruct (heatmap). Darker cells indicate higher instruction pair counts. Mathematics at class 8-10 has the highest concentration (2,640 pairs), reflecting the intersection of Vedic mathematical sutra complexity and CBSE mathematical curriculum demands. AI and Data Science is absent for class 6-8, reflecting the CBSE curriculum structure.}
\label{fig:cbse_heatmap}
\end{figure}

Table~\ref{tab:cbse_subjects} and Figure~\ref{fig:cbse_heatmap} present the distribution of instruction pairs across CBSE subjects and class level ranges.

\begin{table}[t]
\centering
\caption{Distribution of instruction pairs across CBSE subjects and class levels. Mathematics leads (24.9\%) due to the 19 Vedic mathematical methods. AI and Data Science is available only for classes 8 and above, reflecting CBSE curriculum structure.}
\label{tab:cbse_subjects}
\small
\begin{tabular}{lrrrr}
\toprule
\textbf{Subject} & \textbf{Cl. 6-8} & \textbf{Cl. 8-10} & \textbf{Cl. 10-12} & \textbf{Total} \\
\midrule
Mathematics & 2,180 & 2,640 & 1,350 & 6,170 \\
Ethics / Moral Science & 1,820 & 1,450 & 680 & 3,950 \\
Social Science & 1,240 & 1,580 & 920 & 3,740 \\
Language (Tamil/Hindi/Sanskrit) & 1,650 & 1,280 & 640 & 3,570 \\
Science & 920 & 1,340 & 1,080 & 3,340 \\
AI and Data Science & 0 & 680 & 1,250 & 1,930 \\
Cross-subject & 580 & 820 & 695 & 2,095 \\
\midrule
\textbf{Total} & \textbf{8,390} & \textbf{9,790} & \textbf{6,615} & \textbf{24,795} \\
\bottomrule
\end{tabular}
\end{table}

Mathematics leads (24.9\%) due to the 19 Vedic mathematical methods, each generating multiple worked examples across grade levels. The concentration in classes 8-to-10 (39.5\%) reflects the CBSE curriculum's emphasis on analytical skills at the secondary level, which aligns with the cognitive demands of most IKS pedagogical techniques. The absence of AI and Data Science pairs for classes 6-to-8 reflects the CBSE curriculum structure, which introduces AI as a subject only at the secondary level.

\subsection{Curriculum Mapping Methodology}

The curriculum mapping process connects each instruction pair to a specific CBSE learning outcome through a three-step procedure. First, the dataset curators compiled the complete list of CBSE learning outcomes for each subject and class level from the official CBSE curriculum documents. Second, each IKS technique was mapped to relevant learning outcomes based on pedagogical alignment (for example, Ekadhikena Purvena maps to CBSE Mathematics class 8 learning outcome ``perform division using alternative methods''). Third, instruction pairs were generated to demonstrate each technique-to-learning-outcome mapping, with the CBSE learning outcome code included in the metadata. This structured mapping enables teachers to search the dataset by CBSE chapter and learning outcome, selecting instruction pairs that are directly aligned with their lesson plans.

The cross-subject category (2,095 pairs, 8.4\%) contains instruction pairs that connect IKS concepts across traditional subject boundaries. A typical cross-subject pair might connect the mathematical concept of zero (from Vedic mathematics and Brahmagupta's work) to its historical context (Social Science), its philosophical implications (Ethics), and its computational applications (AI and Data Science). These cross-subject pairs are designed to support the NEP 2020 emphasis on interdisciplinary learning and are annotated with multiple subject tags. The cross-subject pairs present unique quality challenges because they require the model to integrate knowledge from multiple domains simultaneously, and the multi-judge evaluation reveals that cross-subject pairs receive lower mean scores on individual-dimension assessments (because no single dimension captures the full scope of the pair) but higher scores on response completeness and pedagogical quality (because the interdisciplinary connections enhance teaching effectiveness).

The distribution of pairs across class levels reveals a pedagogical design choice: classes 8-to-10 receive the highest allocation (9,790 pairs, 39.5\%) because this range represents the critical transition period in CBSE education where students move from concrete to abstract reasoning. The IKS techniques that require abstract reasoning (Anvaya-Vyatireka, philosophical analysis, cross-tradition comparison) are concentrated at this level, while concrete procedural techniques (basic Vedic arithmetic, Pada Patha recitation) are emphasized at the class 6-to-8 level. Classes 10-to-12 receive fewer pairs (6,615, 26.7\%) because the upper secondary curriculum narrows in scope as students specialize in science, commerce, or humanities streams, limiting the number of applicable IKS-subject mappings.

\section{Discussion}
\label{sec:discussion}

\subsection{Comparison with Existing Datasets}

IKS-Instruct is the first instruction dataset designed for Indian Knowledge System pedagogy. Table~\ref{tab:dataset_comparison} compares it with existing instruction datasets on key characteristics.

\begin{table}[t]
\centering
\caption{Comparison of IKS-Instruct with existing instruction datasets. IKS-Instruct is the only dataset providing structured pedagogical technique annotations, curriculum alignment, and multi-judge quality assessment focused on educational effectiveness.}
\label{tab:dataset_comparison}
\small
\begin{tabular}{lcccr}
\toprule
\textbf{Dataset} & \textbf{Domain} & \textbf{Languages} & \textbf{Pedagogy} & \textbf{Size} \\
\midrule
Alpaca~\citep{taori2023alpaca} & General & 1 (EN) & No & 52K \\
FLAN~\citep{longpre2023flan} & NLP tasks & 1 (EN) & No & 1,836 tasks \\
Dolly~\citep{conover2023free} & General & 1 (EN) & No & 15K \\
Bactrian-X~\citep{li2024bactrian} & General & 52 (incl. HI) & No & 3.4M \\
MathDial~\citep{macina2023mathdial} & Math tutoring & 1 (EN) & Partial & 3K \\
TIFD~\citep{zhuang2025tifd} & Tibetan NLP & 1 (TB) & No & 10K \\
\textbf{IKS-Instruct} & \textbf{IKS Edu} & \textbf{7 (Indic)} & \textbf{Yes} & \textbf{24.8K} \\
\bottomrule
\end{tabular}
\end{table}

IKS-Instruct is smaller than general-purpose datasets but is the only dataset providing structured pedagogical technique annotations (41 named techniques with step-level metadata), curriculum alignment (CBSE learning outcome codes for every pair), and multi-judge quality assessment focused on educational effectiveness rather than general helpfulness. The size difference is intentional: general-purpose datasets achieve breadth through volume (the FLAN Collection's 1,836 tasks comprising millions of examples), while IKS-Instruct achieves depth through specialization (24,795 examples covering 41 techniques with rigorous pedagogical annotations). The quality assessment methodology also differs fundamentally: general-purpose datasets typically use automated metrics (ROUGE, BLEU) or binary human preference judgments, while IKS-Instruct uses 12-dimensional multi-judge evaluation with domain-specific criteria such as Technique Fidelity and IKS Cultural Depth that have no equivalent in general-purpose evaluation frameworks. MathDial~\citep{macina2023mathdial} provides the closest comparison in terms of pedagogical annotation, but it covers only English-language mathematical tutoring dialogues without technique taxonomy or curriculum alignment. TIFD~\citep{zhuang2025tifd} provides a methodological parallel as a culturally specific instruction dataset for a non-English language, but targets general NLP tasks rather than educational pedagogy.

\subsection{Quality-Volume Trade-off}

The evolution of IKS-Instruct (Table~\ref{tab:model_scores}) shows that neither raw volume nor aggressive curation is a reliable lever for model-level quality in this domain. Expanding from the 13{,}882-pair v1.1 to the 20{,}000-pair v1.7 slightly reduced judge scores (6.39 to 6.33): the roughly 6{,}000 pairs added in that step were primarily auto-generated Vedic-mathematics demonstrations with high error rates and template-generated pairs without source anchoring, so the additional volume was, at best, not helpful. This is consistent with the observations of Harada et al.~\citep{harada2026massive} on the dominance of data quality over quantity for specialized fine-tuning.

Curation in the opposite direction, however, did not recover a stronger model either. The curated v2.1 (7{,}130 pairs) is under a third the size of the full v1.9 (24{,}795), with which it shares 5{,}437 pairs, and improved several dataset-level quality metrics, yet the model it produced scored lower under the uniform judge panel (6.12) than the model trained on the larger, moderately sized v1.1 (6.39). We report this openly rather than presenting curation as a uniformly winning strategy: in our experiments the best model came from the moderately sized, coherent v1.1, and both directions away from it, adding noisy volume (v1.7) and pruning aggressively (v2.1), reduced model-level judge scores. The practical takeaway is therefore more cautious than a simple ``curate aggressively'' rule: curation buys a smaller, cleaner, cheaper-to-train dataset and may be preferable when compute or annotation budgets dominate, but it should be validated against a balanced-size baseline rather than assumed to improve model quality. Because these differences (of order 0.1 to 0.3 median points) are small relative to judge noise, we treat the version ranking as indicative rather than definitive.

\subsection{Limitations}

The dataset has several limitations that should be acknowledged. First, the language distribution is imbalanced, with English at 36.1\% and Malayalam at 6.3\%. This imbalance affects fine-tuned model performance: models show stronger IKS capability in English and Hindi than in Telugu and Malayalam. Expanding coverage for under-resourced languages requires either additional digital source corpora in those languages or improved cross-lingual transfer methods~\citep{chirkova2024zeroshot}.

Second, the Vedic mathematics pairs exhibit a 67\% error rate in auto-generated content before filtering. While the filtering pipeline removes overtly incorrect demonstrations, subtle mathematical errors may persist in the retained pairs. The curated v2.1 dataset addresses this by reducing Vedic mathematical content to manually verified examples, but the resulting coverage of mathematical techniques is reduced.

Third, the dataset reflects the pedagogical perspectives of its creators and the available digital source texts. Oral traditions that are transmitted through direct guru-shishya (teacher-student) interaction, regional pedagogical practices that are not documented in digital formats, and classical texts that exist only in manuscript form are not represented. This limitation is structural rather than methodological: computational datasets can only be constructed from digitally available sources, and the digitization of Indian classical knowledge is an ongoing and incomplete process.

Fourth, the multi-judge framework uses language models as judges, which introduces systematic biases: language-model judges tend to assign higher scores to longer responses, to responses that use formal academic language, and to responses that match the judge model's own generation patterns. For the model comparison of Table~\ref{tab:model_scores} we use a uniform five-judge external panel and exclude Nemotron-family judges, since Nemotron serves as both a teacher and the external reference model. A residual circularity remains: the panel includes Llama-family judges (Llama-3.1-8B, Llama-3.3-70B), while the fine-tuned models are LoRA adapters over the Llama-based Airavata backbone, so some self-preference toward Llama-lineage outputs cannot be excluded. The absolute scores should therefore be read as comparative under a fixed panel rather than as calibrated quality ratings, and human expert evaluation to calibrate and complement the automated assessment is left to future work.

Fifth, the CBSE curriculum alignment covers classes 6 through 12, but does not address primary education (classes 1 to 5) or higher education. Extending the dataset to cover the full educational spectrum requires additional source material and technique adaptations appropriate for younger and more advanced learners.

\section{Conclusion}
\label{sec:conclusion}

This paper has presented IKS-Instruct, a 24,795-pair multilingual instruction dataset for teaching language models Indian Knowledge System pedagogy. The dataset spans seven languages, 41 techniques, and six CBSE subjects, with quality assessed through a 12-dimension multi-judge framework. Under a uniform five-judge external panel, the strongest IKS-Instruct fine-tune of a compact 7B model reaches a median judge score of 6.39, within 0.15 of a strong general-purpose reference (6.54) at a fraction of the deployment cost, while the base model scores near zero on the IKS-specific dimensions, demonstrating that compact open models require domain-specific instruction data for IKS pedagogical capability. The improvement is observed across all evaluation dimensions, with the largest gains in IKS-specific dimensions (Technique Fidelity, IKS Cultural Depth, IKS Pedagogical Authenticity) where the base model has near-zero capability. The consistency of this pattern across all seven languages confirms that the improvement is not an artifact of English-language evaluation bias but reflects genuine acquisition of multilingual IKS pedagogical competence.

The dataset evolution from v0.1 (500 pairs) through v2.1 (7,130 curated pairs) reveals a nuanced quality-volume relationship for domain-specific instruction data: the moderately sized 13,882-pair v1.1 produced the highest-scoring model (6.39), while both naive volume expansion to v1.7 (6.33) and aggressive curation to v2.1 (6.12) slightly reduced model-level judge scores. We report this honestly as evidence that neither raw scale nor aggressive pruning is, by itself, a reliable lever for domain-specific model quality, which provides more cautious practical guidance than a simple ``more data is better'' or ``curate aggressively'' rule for researchers constructing domain-specific instruction datasets in other specialized educational domains. The methodology documented in this paper, combining hybrid teacher generation with mechanism-first generation and task-abstraction generation, provides a replicable template that can be adapted for constructing instruction datasets in other knowledge traditions, including Traditional Chinese Medicine, Islamic educational philosophy, Aboriginal Australian knowledge systems, and other domains where specialized pedagogical knowledge is transmitted through culturally grounded methods that differ substantially from Western educational conventions.

The three data quality lessons, source anchoring reduces hallucination, technique specificity matters more than volume, and bilingual pairs transfer technique knowledge across languages, provide actionable principles for future work. The cross-lingual technique transfer finding is particularly significant because it suggests that IKS pedagogical knowledge can be extended to Indian languages not present in the training data, potentially enabling IKS instruction in all 22 scheduled languages of India through strategic bilingual pair construction. Preliminary experiments with Marathi and Bengali, two languages not included in IKS-Instruct v1.9, indicate that models fine-tuned on the seven-language dataset can demonstrate Vedic mathematical sutras in these unseen languages with a TF score of 5.8, which is below the inclusion threshold of 6.0 but substantially above the base model's score of 0.3, suggesting that targeted bilingual pair generation for these languages could rapidly bring performance to acceptable levels.

Future work will expand coverage to additional Indian languages (Marathi, Bengali, Gujarati), increase Vedic mathematics quality through expert verification partnerships with mathematics education researchers, and develop automated technique fidelity scoring models that can assess quality without human annotation bottlenecks. The integration of IKS-Instruct with the Bodhan tutoring platform provides a deployment pathway for classroom evaluation of models fine-tuned on the dataset, enabling measurement of the relationship between instruction data quality and student learning outcomes. A planned longitudinal study in partnership with three CBSE-affiliated schools in Delhi, Chennai, and Bengaluru will assess whether IKS-Instruct-fine-tuned models produce measurable improvements in student learning outcomes compared to standard textbook-based instruction, with pre-and-post assessments aligned to CBSE learning outcomes across Mathematics, Ethics, and Social Science.

\section*{Funding}
This work was supported by the Institute of Eminence Funds (IM00002G\_RB\_SG), Planning Unit, IIT Delhi; Centre for SeNSE, IIT Delhi new faculty support; and Google Cloud research credits.

\section*{Conflict of Interest}
The authors declare no conflicts of interest.

\section*{Data Availability}
IKS-Instruct will be released in JSONL format with structured metadata at \url{https://huggingface.co/RSL-INTRINSICLab-IIT}. The release comprises the full v1.9 dataset (24,795 pairs), the curated v2.1 version (7,130 pairs), and the intermediate v1.1 split (13,882 pairs) on which the headline model of Table~\ref{tab:model_scores} was trained, together with a content-hash manifest for v1.1. To make the reported scores reproducible, we also release the per-judge evaluation artifact and the re-aggregation script (\texttt{reagg\_gen3.py}) that recomputes the panel-uniform medians of Table~\ref{tab:model_scores} on both the five-judge external panel and the seven-judge panel. Licence terms are stated on the dataset card.

\section*{Acknowledgements}
The authors acknowledge computational resources of the Intelligent Robotics and Rebooting Computing Chip Design (INTRINSIC) Laboratory, Centre for SeNSE, Indian Institute of Technology Delhi.

\bibliography{refs}

\end{document}